\documentclass[10pt,twocolumn,letterpaper]{article}
\usepackage[accsupp]{axessibility} % Improves PDF readability for those with disabilities.

\usepackage[pagenumbers]{cvpr} % To force page numbers, e.g. for an arXiv version

\usepackage{graphicx}
\usepackage{amsmath}
\usepackage{amssymb}
\usepackage{booktabs}

\usepackage[dvipsnames]{xcolor} % Text color
\usepackage{caption}
\usepackage{subcaption}
\usepackage{multirow}
\usepackage{tabularx}
\usepackage{graphicx}
\usepackage{adjustbox}
\usepackage{amssymb}
\usepackage{pifont}
\usepackage{algorithm}
\usepackage{algorithmic}
\usepackage[skip=2pt]{caption}

\newcommand{\Skip}[1]{}
\newcommand{\paragraphTitle}[1]{\vspace{1mm}\noindent{\textbf{#1}\hspace{1mm}}}
\usepackage[pagebackref,breaklinks,colorlinks]{hyperref}

\usepackage[capitalize]{cleveref}
\crefname{section}{Sec.}{Secs.}
\Crefname{section}{Section}{Sections}
\Crefname{table}{Table}{Tables}
\crefname{table}{Tab.}{Tabs.}

\def\confName{CVPR}
\def\confYear{2022}

\begin{document}

\title{Part-based Pseudo Label Refinement for Unsupervised Person Re-identification}
\author{
    \begin{tabular}[t]{cccc} Yoonki Cho & Woo Jae Kim & Seunghoon Hong & Sung-Eui Yoon \end{tabular}\vspace{1.5mm} \\
    KAIST \\
    {\tt\small \{yoonki, wkim97, seunghoon.hong, sungeui\}@kaist.ac.kr}
}
\maketitle

\urlstyle{same}
\begin{abstract}
Unsupervised person re-identification (re-ID) aims at learning discriminative representations for person retrieval from unlabeled data.
Recent techniques accomplish this task by using pseudo-labels, but these labels are inherently noisy and deteriorate the accuracy.
To overcome this problem, several pseudo-label refinement methods have been proposed, but they neglect the fine-grained local context essential for person re-ID.
In this paper, we propose a novel Part-based Pseudo Label Refinement (PPLR) framework that reduces the label noise by employing the complementary relationship between global and part features.
Specifically, we design a cross agreement score as the similarity of $k$-nearest neighbors between feature spaces to exploit the reliable complementary relationship.
Based on the cross agreement, we refine pseudo-labels of global features by ensembling the predictions of part features, which collectively alleviate the noise in global feature clustering.
We further refine pseudo-labels of part features by applying label smoothing according to the suitability of given labels for each part.
Thanks to the reliable complementary information provided by the cross agreement score, our PPLR effectively reduces the influence of noisy labels and learns discriminative representations with rich local contexts.
Extensive experimental results on Market-1501 and MSMT17 demonstrate the effectiveness of the proposed method over the state-of-the-art performance.
The code is available at \url{https://github.com/yoonkicho/PPLR}.
\end{abstract}

\section{Introduction}
\label{sec:1}
    Person re-identification (re-ID) aims to retrieve a person corresponding to a given query across disjoint camera views or different time stamps~\cite{ye2021deep, zheng2016person}.
    Thanks to the discriminative power of deep neural networks, supervised approaches~\cite{li2014deepreid, wei2017glad, li2017person, li2018harmonious} have achieved impressive performance in this task.
    Unfortunately, they require a large amount of labeled data that demands costly annotations, limiting their practicality in large-scale real-world re-ID problems.
    Due to this issue, unsupervised methods that learn the discriminative features for person retrieval from unlabeled data have recently received much attention.

    Prior works on unsupervised person re-ID have utilized pseudo-labels obtained by $k$-nearest neighbor search~\cite{lin2020unsupervised, wang2020unsupervised, yu2019unsupervised} or unsupervised clustering~\cite{lin2019aBottom, fu2019self} for training.
    These approaches alternate a two-stage training scheme: the label generation phase that assigns pseudo-labels and the training phase that trains a model with generated labels.
    Among these approaches, the clustering-based methods~\cite{chen2021ice, ge2020selfpaced} have especially demonstrated their effectiveness with state-of-the-art performance.
    However, inherent noises in pseudo-labels significantly hinder the performance of these unsupervised methods.

    To tackle this problem, many efforts have been made to improve the accuracy of pseudo-labels by performing robust clustering~\cite{ge2020selfpaced, zhai2020ad} or pseudo-label refinement~\cite{lin2020unsupervised, zhang2021refining}.
    Recent techniques~\cite{ge2020mutual, zhai2020multiple} significantly reduce the label noise through the model ensemble in a peer-teaching manner by using predictions from an auxiliary network as refined labels for the target network.
    Nevertheless, training multiple backbones as teacher networks (\textit{e.g.}, dual ResNet in MMT~\cite{ge2020mutual}, and single DenseNet, ResNet, and Inception-v3 in MEB-Net~\cite{zhai2020multiple}) requires high computational costs.
    Furthermore, labels refined by these methods consider only global features and neglect the fine-grained clues essential to person re-ID, leading to insufficient performance.

    To address the aforementioned problems, we propose \textit{Part-based Pseudo Label Refinement (PPLR)}, a novel unsupervised re-ID framework that effectively handles the label noise using part features in a self-teaching manner.
    Several studies \cite{sun2018beyond, zheng2019pyramidal} demonstrate that the fine-grained information from part features improves the re-ID performance.
    Our key idea is that this fine-grained information can provide not only useful cues for better representation learning but also robustness against label noises.
    In contrast to the global-shape information that has large variations due to significant changes in poses and viewpoints, part features can capture the local-texture information that provides a more crucial clue to re-identifying a person~\cite{zheng2019joint}.

    We argue that the complementary relationship between the global and part features can be used to refine the label noise in each of their feature spaces.
    However, some of the global and part features from the same image capture very different semantic information, and using the complementary relationship na\"ively can result in noisy and even incorrect information.
    For instance, images may contain irrelevant parts (\textit{e.g.,} occlusions or backgrounds) that provide unreliable complementary information, and it is desirable to exclude them from the training.
    Therefore, it is essential to identify whether the information of global and part features are reliable with each other to properly exploit their complementary relationship.
    To address this issue, we design a cross agreement score based on the similarity between the $k$-nearest neighbors of global and part features.
    Based on the cross agreement, we propose two pseudo-label refinement methods -- \textit{part-guided label refinement (PGLR)} and \textit{agreement-aware label smoothing (AALS)}.
    PGLR refines the pseudo-labels of global features by aggregating the predictions of part features, guiding the global features to learn from rich local contexts.
    AALS refines the pseudo-labels of part features by smoothing the label distributions, thus calibrating the predictions of part features.
    
    Our contributions can be summarized as follows:
    \begin{itemize}
    \vspace{-2.5mm}
    \item We propose a part-based pseudo-label refinement framework that operates in a self-ensemble manner without auxiliary networks.
          To the best of our knowledge, this is the first work to handle the label noise using the part feature information for person re-ID.
    \vspace{-2.5mm}
    \item We design a cross agreement score to capture reliable complementary information, which is computed by the similarity between the $k$-nearest neighbors of the global and part features.
    \vspace{-2.5mm}
    \item Extensive experimental results with superior performance against the state-of-the-art methods demonstrate the effectiveness of the proposed method.
    \end{itemize}

\section{Related Work} \label{sec:2}

\noindent\textbf{Learning with noisy labels.}{
    Due to the difficulty of obtaining high-quality labels in many real-world scenarios, much attention has been given to robust training with noisy labels~\cite{song2020learning}.
    Loss adjustment approaches utilize the loss correction technique through the noise transition matrix~\cite{xiao2015learning, vahdat2017toward, patrini2017making, hendrycks2018using} or the sample re-weighting scheme based on the reliability of a given label~\cite{chang2017active, ren2018learning, shu2019meta} to reduce the influence of noisy labels.
    However, these methods require a certain number of clean labels to estimate the degree of noise and are not applicable for unsupervised person re-ID, where the pseudo-labels are extremely noisy at the beginning of training.
    There also have been attempts to design a robust loss function against the label noise.
    Ghosh \textit{et al.}~\cite{ghosh2017robust} demonstrate that mean absolute error (MAE) loss is theoretically robust to noisy labels.
    Generalized cross-entropy (GCE)~\cite{zhang2018generalized} loss was proposed to achieve both the robustness of MAE and better convergence of the cross-entropy loss.
    Wang \textit{et al.}~\cite{wang2019symmetric} propose symmetric cross-entropy (SCE) loss which boosts the noise tolerance with the reverse cross-entropy loss.
    These loss functions, however, are designed for simple image classification tasks and are not suitable for open-set person re-ID tasks.
}

\paragraphTitle{Part-based approaches for person re-ID.}{
    Fine-grained information on human body parts is an essential clue to distinguish people, and recent studies~\cite{zhao2017deeply, wei2017glad, su2017pose} leverage part features and show state-of-the-art performance.
    There have been many efforts to learn more discriminative part features through human parsing~\cite{guo2019beyond, kalayeh2018human}, attention mechanism~\cite{li2018harmonious, si2018dual, yang2019attention}, pose estimation~\cite{suh2018part, miao2019pose}, and multiple granularities~\cite{zheng2019pyramidal, wang2018learning}.
    Despite the remarkable achievements of part-based supervised approaches, there have been only a few attempts to utilize part features for unsupervised person re-ID.
    For domain adaptive re-ID, PAUL~\cite{yang2019patch} extracts part features using a spatial transformer network~\cite{jaderberg2015spatial} for patch-based discriminative learning.
    Recent methods~\cite{lin2020unsupervised, fu2019self} utilize part features to exploit robust feature similarity for accurate pseudo-labels.
    Contrary to the above methods, our work utilizes part features to reduce the label noise of global feature clustering by providing fine-grained information.
}

\paragraphTitle{Unsupervised approaches for person re-ID.} {
    The existing unsupervised approaches can be divided into unsupervised domain adaptation (UDA) and unsupervised learning depending on whether an external labeled source domain data is used.
    Several UDA methods reduce the domain gap between the source and target datasets through feature distribution alignment~\cite{lin2018multi, wang2018transferable} and image style transfers~\cite{wei2018person, deng2018image, Zhong_2018_ECCV}.
    In recent years, both UDA and unsupervised learning methods leverage pseudo-labels assigned by clustering~\cite{lin2019aBottom, zeng2020hierarchical, song2020unsupervised} or nearest neighbor search~\cite{wang2020unsupervised, yu2019unsupervised, zhong2019invariance}.
    Recent clustering-based methods~\cite{ge2020selfpaced, wang2021camera, chen2021ice} apply the contrastive learning scheme with cluster proxies and show impressive achievements.
    However, the inherent label noise in pseudo-labels degrades the performance, and many recent studies tackle this essential problem.
    SSL~\cite{lin2020unsupervised} softens pseudo-labels by measuring the robust similarity with additional information such as camera labels.
    MMT~\cite{ge2020mutual} and MEBNet~\cite{zhai2020multiple} refine pseudo-labels using the predictions of auxiliary teacher networks in a mutual learning manner.
    RLCC~\cite{zhang2021refining} utilizes a label propagation scheme based on the clustering consensus matrix to reduce the label noise.
    Unlike these label refinement methods that only consider the global context, our work employs fine-grained information from the part features to refine the labels more effectively.
}

\section{Method} 
\label{sec:3}
    \begin{figure*}[t]
        \centering
        \includegraphics[width=15cm]{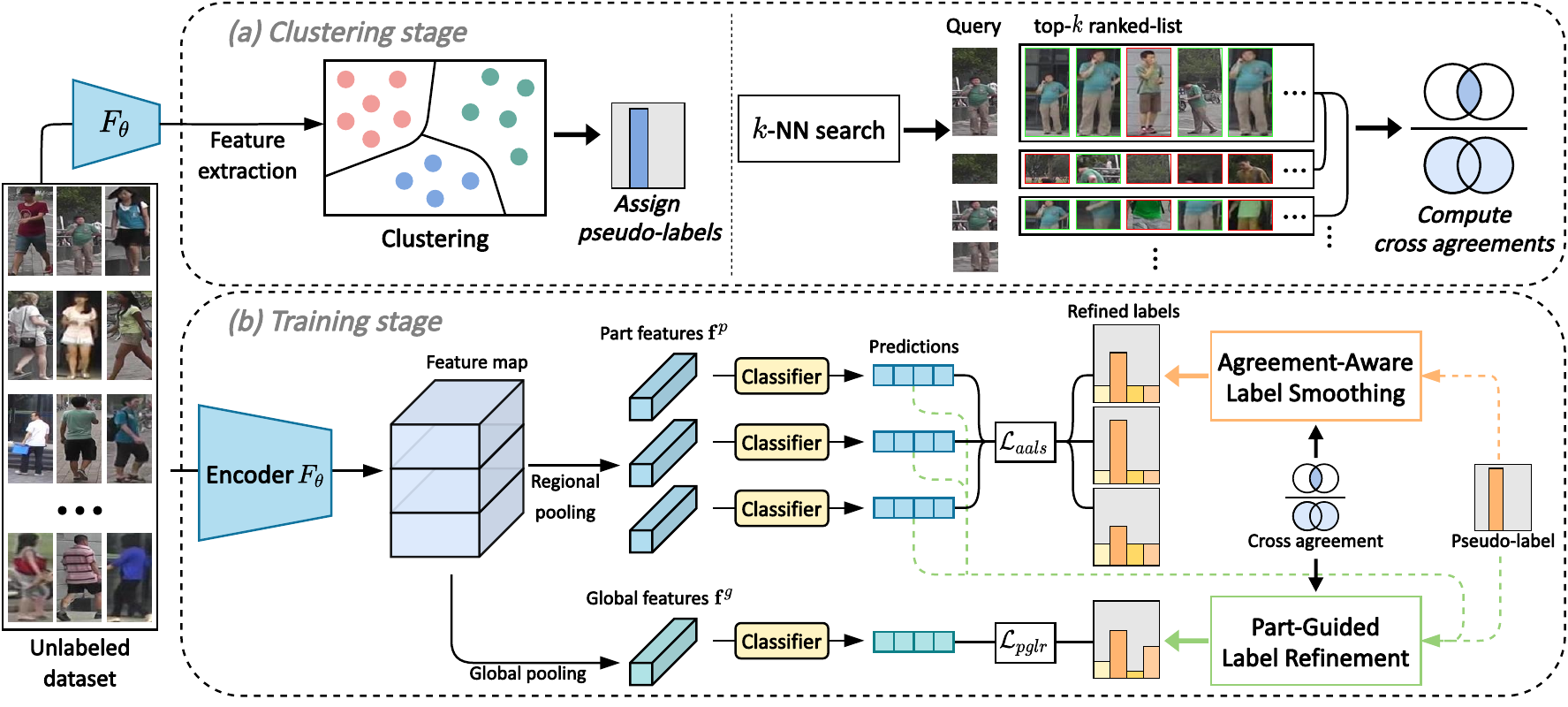}
        \caption{ 
            The illustration of PPLR.
            Our method alternates the clustering stage and the training stage.
            (a) In the clustering stage, we assign pseudo-labels by clustering the global features on the unlabeled dataset.
            We then perform a $k$-nearest neighbor search on each feature space and compute the cross agreement score based on the similarity between the top-$k$ ranked lists of the global and part features.
            (b) In the training stage, we train the model with refined pseudo-labels based on the cross agreement score.
            We smooth the labels of part features according to the cross agreement score of each part and refine the labels of global features by aggregating the part features' predictions.
        }
    \label{fig:overview}
    \vspace{-5mm}
    \end{figure*}
    
    We propose a {Part-based Pseudo Label Refinement (PPLR)} framework that exploits the complementary relationship between the global and part features to tackle the label noise problem.
    Following the existing clustering-based methods~\cite{lin2019aBottom, zhang2021refining, zeng2020hierarchical}, our method alternates the clustering stage and the training stage.
    In the clustering stage, we extract global and part features and assign pseudo-labels through global feature clustering.
    We then compute a cross agreement score for each sample based on the similarity between $k$-nearest neighbors of global and part features.
    In the training stage, we mitigate the label noise using the proposed pseudo-label refinement methods based on the cross agreement: {agreement-aware label smoothing (AALS)} for part features and {part-guided label refinement (PGLR)} for global features.
    The features from the trained model are then used in the next clustering stage to update the pseudo-labels.
    The overall framework is illustrated in Fig.~\ref{fig:overview}.

\subsection{Part-based Unsupervised re-ID Framework}
\label{ssec:3-1}
    We first present a part-based unsupervised person re-ID framework that utilizes fine-grained information of the part features.
    Contrary to most existing unsupervised approaches that exploit only the global feature, we use both the global and part features to represent an image.

    Formally, let ${\mathcal{D} = \{x_i\}_{i=1}^{N_\mathcal{D}}}$ denote the unlabeled training dataset, where $x_i$ is an image and $N_\mathcal{D}$ is the number of images.
    Our model first extracts the shared representation ${F}_\theta(x_i)\in\mathbb{R}^{C \times H \times W}$, where $C$, $H$, and $W$ are sizes of the channel, height, and width of the feature map, respectively.
    Given this feature map, the global feature $\mathbf{f}^g_i$ is obtained by applying global average pooling over the feature map, while the part features $\{{\mathbf{f}^{p_n}_i}\}^{N_p}_{n=1}$ are obtained by dividing the feature map into $N_p$ uniformly partitioned regions $\mathbb{R}^{C \times \frac{H}{N_p} \times W}$ and applying average pooling on each region.
    
    To learn these representations without a label, we simulate the pseudo-labels based on clustering results.
    Following the part-based approaches~\cite{sun2018beyond, wang2018learning, zheng2019pyramidal} in the literature, we adopt the standard protocol where both the global and part features share the same pseudo-labels.
    We perform DBSCAN clustering \cite{ester1996density} on the global feature set $\{\mathbf{f}^g_i\}_{i=1}^{N_\mathcal{D}}$ and use the cluster assignment as pseudo-labels.
    We denote the pseudo-label for the image $x_i$ as $y_i \in \mathbb{R}^K$, which is the one-hot encoding of the hard assignment with $K$ clusters.

    The pseudo-labels are then used to train the global and part features for person identification.
    For global features, we compute the cross-entropy loss by:
    \begin{equation}
            \mathcal{L}_{gce} = -\sum_{i=1}^{N_\mathcal{D}}{y_i \cdot \log({q^g_i})},
            \label{eq:gce}
    \end{equation}
    where ${q^g_i} = h_{\phi_g}(\mathbf{f}^g_i) \in \mathbb{R}^K$ is the prediction vector by the global feature, and $h_{\phi_g}(\cdot)$ is the global feature classifier consisting of a fully connected layer and a softmax function.
    Similarly, we train the part features using the cross-entropy loss by:
    \begin{equation}
            \mathcal{L}_{pce} = -\frac{1}{N_p}\sum_{i=1}^{N_\mathcal{D}}\sum_{n=1}^{N_p}{y_i \cdot \log({q^{p_n}_i}}),
            \label{eq:pce}
    \end{equation}
    where $q_i^{p_n} = h_{\phi_{p_n}}(\mathbf{f}^{p_n}_i) \in \mathbb{R}^K$ indicates the prediction vector by the $n$-th part feature space $p_n$, and $h_{\phi_{p_n}}$ is the classifier for the part feature space $p_n$.
    We additionally utilize the softmax-triplet loss defined by:
    \begin{equation}
            \mathcal{L}_{triplet} = -\sum_{i=1}^{N_\mathcal{D}}
            {\log
            \left(
            {\frac {e^{\lVert \mathbf{f}_i^g - \mathbf{f}_{i,n}^g\rVert}}
            {e^{\lVert \mathbf{f}_i^g - \mathbf{f}_{i,p}^g\rVert}+e^{\lVert \mathbf{f}_i^g - \mathbf{f}_{i,n}^g\rVert}}}
            \right)},
            \label{eq:tripelt}
    \end{equation}
    where $\lVert \cdot \rVert$ denotes the $L_2$-norm, and the subscripts $(i,p)$ and $(i,n)$ respectively denote the hardest positive and negative samples of the image $x_i$ in a mini-batch obtained by the hard-batch triplet selection~\cite{hermans2017defense}.
    Following the recent studies~\cite{chen2021ice, xuan2021intra} that utilize camera labels to improve the discriminability across camera views, we can optionally employ a camera-aware proxy~\cite{wang2021camera} if the camera labels are available.
    We compute the camera-aware proxy $\mathbf{c}_{(a,b)}$ as the centroid of the features that have the same camera label $a$ and belong to the same cluster $b$.
    We then compute the inter-camera contrastive loss~\cite{wang2021camera} as:
    \begin{equation}
            \mathcal{L}_{cam} = -\sum_{i=1}^{N_\mathcal{D}}
            \frac{1}{\lvert P_i \rvert} 
            \sum_{j \in \mathcal{P}_i}
            {\log{   
            \frac{\exp(\mathbf{c}^\top_j \mathbf{f}^g_i / \tau )}
            {\sum\limits_{k \in \mathcal{P}_i\cup \mathcal{Q}_i} \exp(\mathbf{c}^\top_k \mathbf{f}^g_i / \tau )}
            }},
            \label{eq:cam}
    \end{equation}
    where $\mathcal{P}_i$ and $\mathcal{Q}_i$ are the index sets of the positive and hard negative camera-aware proxies{\footnote{More details can be found in the appendix.}} for $\mathbf{f}^g_i$, and $\tau$ is the temperature parameter.
    This loss pulls together the proxies that are within the same cluster but in different cameras, reducing the intra-class variance caused by disjoint camera views.
    The training objective is then given by:
    \begin{equation}
            \mathcal{L} = \mathcal{L}_{gce} + \mathcal{L}_{pce} + \mathcal{L}_{triplet} + \lambda_{cam}\mathcal{L}_{cam},
            \label{eq:baseline}
    \end{equation}
    where $\lambda_{cam}$ is the weight parameter that controls the importance of the inter-camera contrastive loss.
    
    Ideally, the model can learn both the holistic and local features by sharing the common representation $F_\theta(\cdot)$.
    However, its performance is inherently bounded by the quality of the pseudo-label $y_i$, which is significantly noisy in practice.
    In the following sections, we propose a method to refine such noisy labels for properly representing both features.
    
    \begin{figure}[t]
        \centering
        \includegraphics[width=0.95\columnwidth]{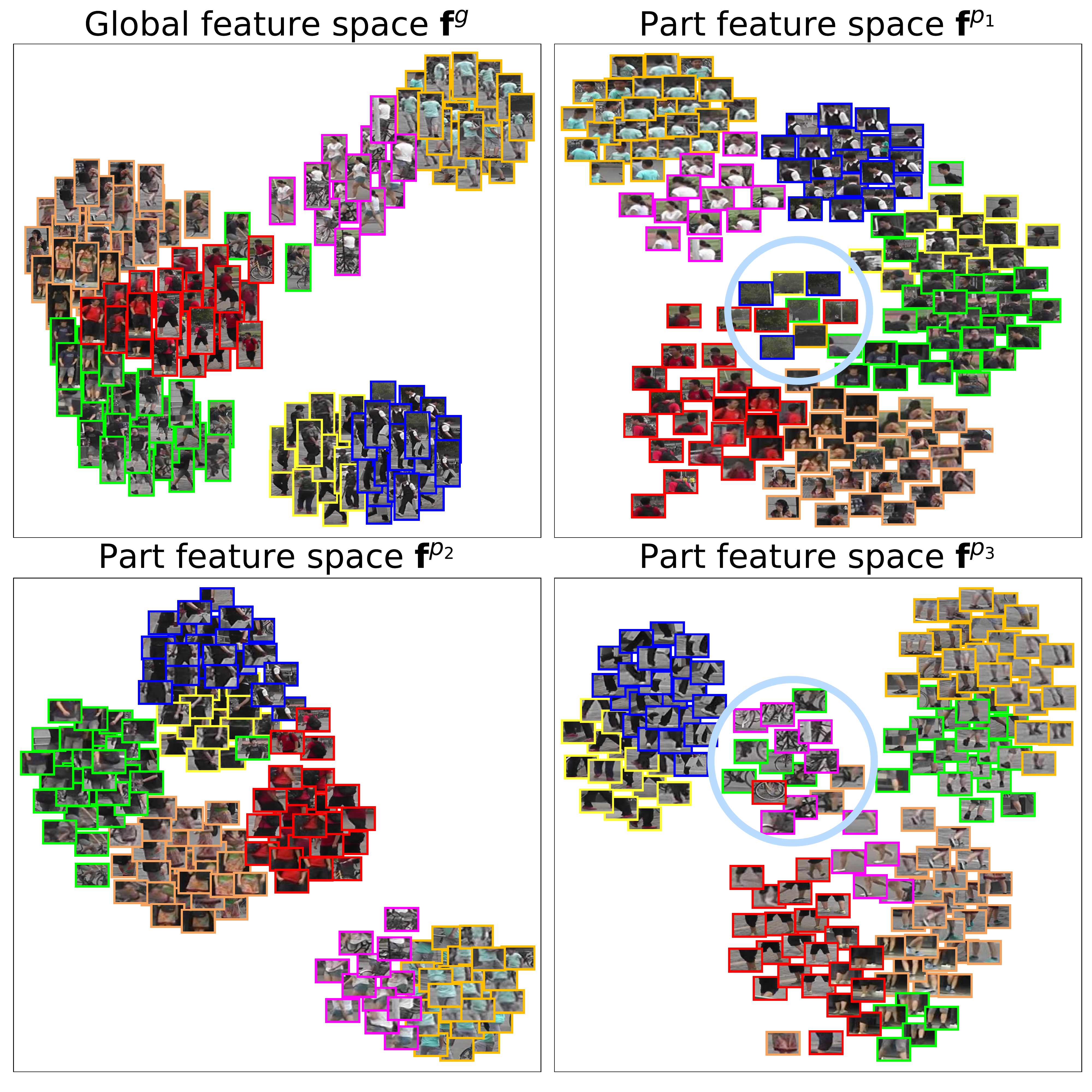}
        \caption{ 
            The t-SNE~\cite{van2008visualizing} visualization of each feature space on Market-1501 at an early training epoch.
            Different bounding box colors represent different IDs.
            Each feature space shows a different feature distribution with different semantic parts of a person, and some feature information can be unreliable.
            For instance, less discriminative parts denoted by a circle provide irrelevant information to their counterparts in other feature spaces and vice-versa.
            Our cross agreement score identifies such noisy information by comparing the $k$-nearest neighbors between feature spaces.
        }
        \label{fig:tsne-each-space}
        \vspace{-5mm}
    \end{figure} 

\subsection{Cross Agreement} \label{ssec:3-2}
    Contrary to our basic framework, PPLR trains the model with refined pseudo-labels that consider the complementary relationship between global and part features.
    Nevertheless, there exists unreliable complementary information due to differences in feature similarity structures between global and part features.
    As shown in Fig.~\ref{fig:tsne-each-space}, some part features contain information irrelevant to a person and are not suitable for refining pseudo-labels of global features.
    Furthermore, global features consider only the global context and sometimes neglect information relevant to part features.
    Therefore, identifying whether the given complementary information is reliable is an essential task for our method.
    
    To address this issue, we design a cross agreement score that captures how reciprocally similar the $k$-nearest neighbors of global and part features are.
    We define the cross agreement score as the Jaccard similarity between the $k$-nearest neighbors of the global and part features.
    We first perform a $k$-nearest neighbor search on the global and each of the part feature spaces independently to produce $(1+N_p)$ ranked lists on each image.
    We then compute the cross agreement score between the global feature space $g$ and the $n$-th part feature space $p_n$ for the image $x_i$ by:
    \begin{equation}
            \mathcal{C}_i(g, p_n) = 
            \frac
            {\lvert\mathcal{R}_{i}(g, k) \cap \mathcal{R}_{i}(p_n, k)\rvert}
            {\lvert\mathcal{R}_{i}(g, k) \cup \mathcal{R}_{i}(p_n, k)\rvert}
            \in [0,1],
            \label{eq:agreement}
    \end{equation}
    where $\mathcal{R}_{i}(g, k)$ and $\mathcal{R}_{i}(p_n, k)$ are the sets of indices for the top-$k$ samples in the ranked list computed by $\mathbf{f}^g_i$ and $\mathbf{f}^{p_n}_i$, respectively, and $\lvert \cdot \rvert$ is the cardinality of a set. 
    
    Intuitively, a high cross agreement score $\mathcal{C}_i(g, p_n)$ implies that the feature spaces of $g$ and $p_n$ have highly correlated feature similarity structure around the data point $i$ and provide reliable complementary information. 
    On the other hand, a low $\mathcal{C}_i(g, p_n)$ implies that the global and part features are not very correlated, meaning that they can provide unreliable information to each other.
    Our cross agreement score is designed in the same spirit as recent re-ranking techniques~\cite{sarfraz2018pose, zhong2017re, yang2019efficient, iscen2017efficient} that utilize a reciprocity check of k-nearest neighbors to handle the similarity noise in the affinity matrix to improve retrieval performance.
    
\subsection{Pseudo Label Refinement} \label{ssec:3-3}
    Based on the cross agreement scores, we alleviate the label noise by considering (1) whether the pseudo-labels by global feature clustering are suitable for each part feature and (2) whether the predictions of part features are appropriate for refining pseudo-labels of global features.

\paragraphTitle{Agreement-aware label smoothing.}
    Learning all part features with the same global pseudo-label that neglects the local context of parts can be detrimental to the model training.
    For instance, some parts contain cues irrelevant to a person (\textit{e.g.}, occlusions), and it is desirable to exclude them from the training.
    To address this issue, we utilize a label smoothing~\cite{szegedy2016rethinking, lukasik2020does} to refine the pseudo-label of each part depending on the corresponding cross agreement score.

    Given the pseudo-label $y_i$ of the image $x_i$, the label smoothing for the part feature $\mathbf{f}^{p_n}_i$ is formulated as:
    \begin{equation}
        \tilde{y}_i^{p_n} =  \alpha_i^{p_n} \, y_i + (1- \alpha_i^{p_n}) \, u,
    \label{eq:ls}
    \end{equation}
    where $u$ is a uniform vector, and $\alpha_i^{p_n}$ is a weight determining the strength of label smoothing.
    Contrary to conventional label smoothing that employs a constant weight for $\alpha_i^{p_n}$,
    we dynamically adjust the weight for each part using the cross agreement score (\textit{i.e.}, $\alpha_i^{p_n}= \mathcal{C}_i(g, p_n)$) 
    that reflects the reliability of the global clustering result for each part.
    We then plug the refined pseudo-labels $\tilde{y}_i^{p_n}$ to Eq.~\eqref{eq:pce}, and the cross-entropy loss is reformulated with Kullback-Leibler (KL) divergence~\cite{pereyra2017regularizing} by:
    \begin{multline}
            \mathcal{L}_{aals} = \frac{1}{N_p} \sum_{i=1}^{N_\mathcal{D}}\sum_{n=1}^{N_p}(\alpha_i^{p_n} \textit{H}({y}_i, q_i^{p_n}) \\ + (1-\alpha_i^{p_n})\textit{D}_\mathrm{KL}(u \parallel q_i^{p_n})),
    \label{eq:aals}
    \end{multline}
    where $\textit{H}(\cdot,\cdot)$ and $\textit{D}_\mathrm{KL}(\cdot \! \parallel \! \cdot)$ are cross-entropy and KL divergence, respectively, and two terms are balanced by $\alpha_i^{p_n}$ with the value of the cross agreement score $\mathcal{C}_i(g, p_n)$.
    
    In Eq.~\eqref{eq:aals}, the former term drives the prediction to high confidence close to $y_i$, and the latter term encourages the prediction to collapse into a uniform vector. 
    By scaling the two opposite terms with the cross agreement scores, we calibrate the prediction of part features according to the reliability of pseudo-labels for each part.
    
\paragraphTitle{Part-guided label refinement.}
    We propose a part-guided label refinement that generates refined labels for global features using the predictions by part features.
    The part feature information with rich local contexts can be used to handle the label noise in global feature clustering, which often neglects fine-grained information. 
    However, since less discriminative parts can provide misleading information, we aggregate the predictions of part features with different weights depending on each cross agreement score, thus refining the labels with more reliable information.
    
    Specifically, we generate the part-guided refined label, $\tilde{y}_i^g$, as a pseudo-label for the global feature by:
    \begin{equation}
        \tilde{y}_i^g = \beta y_i + (1-\beta) \sum_{n=1}^{N_p} w^{p_n}_i q^{p_n}_i,
    \label{eq:pglr}
    \end{equation}
    where $w^{p_n}_i = \frac{\exp(\mathcal{C}_i(g, p_n))}{\sum_{k}\exp(\mathcal{C}_i(g, p_k))}$ and $q^{p_n}_i$ are the ensemble weight and the prediction vector of the part feature $\mathbf{f}^{p_n}_i$, respectively.
    $\beta \in [0, 1]$ is the weighting parameter controlling the ratio of the one-hot pseudo-label and the ensembled prediction.
    Contrary to the global feature that only captures holistic characteristics of a person, the part-guided refined label in Eq.~\eqref{eq:pglr} additionally considers the fine-grained predictions from the local parts in proportion to their reliability captured by the cross agreement score.
    The refined labels $\tilde{y}_i^g$ are then plugged to Eq.~\eqref{eq:gce} to train the global feature by:
    \begin{equation}
            \mathcal{L}_{pglr} = -\sum_{i=1}^{N_\mathcal{D}}{\tilde{y}_i^g \cdot \log({q_i^g}}).
            \label{eq:pglr-ce}
    \end{equation}
    
    With the part-guided refined labels, global features learn from the ensembled part predictions with rich fine-grained information that is neglected in previous methods. 
    Furthermore, unlike previous studies~\cite{zhai2020multiple, ge2020mutual} that refine pseudo-labels using an auxiliary teacher network, a part-guided label refinement is a self-teaching method without an additional network, being computationally efficient.

\paragraphTitle{Overall training objective.}
    The overall loss function of PPLR is then:
    \begin{equation}
        \mathcal{L}_\mathrm{PPLR} = \mathcal{L}_{aals} + \mathcal{L}_{pglr} + \mathcal{L}_{triplet} + \lambda_{cam}\mathcal{L}_{cam}.
    \label{eq:pplr}
    \end{equation}
    
    Our method effectively reduces the influence of noisy labels in two ways. 
    The part features with low cross agreements are trained by pseudo-labels close to a uniform distribution by Eq.~\eqref{eq:aals}, 
    and the global features trained by the part-guided refined labels capture reliable fine-grained information from the part features by Eq.~\eqref{eq:pglr}.
    Also, when all part predictions have low cross agreement scores, the ensembled prediction in the part-guided refined label eventually collapses to a uniform vector due to the strong label smoothing effect in all parts, thus providing meaningless training signals.
    It allows us to weaken the impact of noisy pseudo-labels, resulting in better representation learning.

\section{Experiments} \label{sec:4}
\subsection{Datasets and Evaluation Protocols}
    We evaluate the proposed method on two person re-ID datasets and a vehicle re-ID dataset -- Market-1501~\cite{zheng2015scalable}, MSMT17~\cite{wei2018person}, and VeRi-776~\cite{liu2016deep}.
    Market-1501 contains 32,688 images of 1,501 person identities from 6 non-overlapping camera views.
    It is split into 12,936 training images of 751 identities and 19,732 testing images of 750 identities.
    MSMT17 is a more challenging dataset consisting of 126,441 images of 4,101 person identities captured from 15 different cameras. 
    It is split into 32,621 training images of 1,041 identities and 93,820 testing images of 3,060 identities.
    VeRi-776 contains 51,035 images of 776 vehicles from 20 camera views.
    It is split into 37,778 training images of 576 vehicles and 13,257 testing images of 200 vehicles.
    We adopt mean average precision (mAP) and cumulative matching characteristic (CMC) Rank-1, Rank-5, Rank-10 accuracies to evaluate performances.

\paragraphTitle{Implementation Details.}{
    We adopt ResNet-50~\cite{he2016deep} pre-trained on ImageNet~\cite{deng2009imagenet} as the backbone.
    We remove all layers after layer-4 and add average pooling layers followed by fully-connected classifiers with BNNeck~\cite{luo2019strong}.
    During testing, we use only global features for retrieval.
    The person and vehicle images are resized to $384\times128$ and $256\times256$, respectively.
    Random flipping, cropping, and erasing \cite{zhong2020random} are used for data augmentation.
    The mini-batch size is 64 consisting of 16 pseudo-classes and 4 images for each class.
    Adam \cite{kingma2015adam} with weight decay of $5\times{10}^{-4}$ is adopted for training.
    The initial learning rate is set to $3.5\times{10}^{-4}$ and is decreased by a factor of 10 after every 20 epochs.
    We train for a total of 50 epochs, and each epoch contains 400 iterations.
    We employ DBSCAN \cite{ester1996density} based on Jaccard distance with $k$-reciprocal encoding \cite{zhong2017re} for clustering.
    We empirically set the number of parts $N_p$ to 3, the weighting parameter $\beta$ to 0.5, and the parameter $k$ of cross agreement scores to 20.
    Following CAP~\cite{wang2021camera}, we set $\tau=0.07$, $\lambda_{cam}=0.5$, and the number of hard negative proxies to 50.
    }
    
    \begin{table}[t]
    	\centering	
    	\begin{adjustbox}{max width=\columnwidth}
    	\footnotesize
        \begin{tabular}{l|cc|cc}
        \hline
        \multicolumn{1}{c|}{\multirow{2}{*}{Method}} & \multicolumn{2}{c|}{Market-1501} & \multicolumn{2}{c}{MSMT17} \\ \cline{2-5} 
         & mAP & Rank-1 & mAP & Rank-1         \\ \hline
        Baseline1 (w/o $\mathcal{L}_{cam}$)  & 73.5 & 88.5 & 25.1 & 51.2 \\
        + $\mathcal{L}_{aals}$ & 78.9 & 90.9 & 28.8 & 56.1 \\
        + $\mathcal{L}_{pglr}$ & 77.8 & 90.7 & 29.3 & 55.6 \\
        + $\mathcal{L}_{aals}$ + $\mathcal{L}_{pglr}$ & \textbf{81.5} & \textbf{92.8} & \textbf{31.4} & \textbf{61.1} \\ \hline
        Baseline2 (w/ $\mathcal{L}_{cam}$) & 77.9 & 91.2 & 36.8 & 67.6 \\
        + $\mathcal{L}_{aals}$ & 82.4 & 93.2 & 40.5 & 71.2 \\
        + $\mathcal{L}_{pglr}$ & 81.5 & 93.0 & 40.4 & 70.9 \\
        + $\mathcal{L}_{aals}$ + $\mathcal{L}_{pglr}$ & \textbf{84.4} & \textbf{94.3} & \textbf{42.2} & \textbf{73.3} \\ \hline
    	\end{tabular}
    	\end{adjustbox}
    	\caption{Ablation study on individual components of PPLR.}
    	\label{tab:ablation_study}
    \vspace{-5mm}
    \end{table}

\subsection{Ablation Study}
    To analyze the effectiveness of our method, we conduct extensive experiments on Market-1501 and MSMT17.
    Since camera labels are not always available in practice, we adopt the part-based unsupervised re-ID framework (Sec.~\ref{ssec:3-1}) with and without the inter-camera contrastive loss $\mathcal{L}_{cam}$ as the baselines.
    We evaluate the effectiveness of the proposed pseudo-label refinement methods -- agreement-aware label smoothing $\mathcal{L}_{aals}$ and part-guided label refinement $\mathcal{L}_{pglr}$. 
    The experimental results on each setting are reported in Table~\ref{tab:ablation_study}.
    As shown in the table, each label refinement significantly boosts performances on both settings.
    We especially obtain remarkable performance gains when we combine AALS and PGLR. For instance, our method improves mAP of the baselines with and without $\mathcal{L}_{cam}$ by 6.5\% and 8.0\%, respectively, on Market-1501.
    
    \begin{table}[t]
    	\centering	
    	\begin{adjustbox}{max width=\columnwidth}
    	\footnotesize
        \begin{tabular}{l|cc|cc}
        \hline
        \multicolumn{1}{c|}{\multirow{2}{*}{Method}} & \multicolumn{2}{c|}{Market-1501} & \multicolumn{2}{c}{MSMT17} \\ \cline{2-5} 
         & mAP & Rank-1 & mAP & Rank-1 \\ \hline
        PPLR & \textbf{81.5} & \textbf{92.8} & \textbf{31.4} & \textbf{61.1}\\ \hline
        PPLR w/o $\mathcal{L}_{aals}$ & 77.8 & 90.7 & 29.3 & 55.6 \\
        + \textit{Part-to-part label refinement} & 78.6 & 91.1 & 27.9 & 54.9 \\
        + \textit{Vanilla label smoothing} & 79.1 & 91.7 & 29.8 & 57.2 \\ \hline
        PPLR w/o $\mathcal{L}_{pglr}$  & 78.9 & 90.9 & 28.8 & 56.1 \\
        + \textit{Mean-teaching} & 79.0 & 91.0 & 28.9 & 55.7 \\
        + $\mathcal{L}_{pglr}$ w/o \textit{cross agreement} & 80.1 & 91.5 & 30.1 & 58.1 \\ \hline
    	\end{tabular}
    	\end{adjustbox}
    	\caption{Ablation study on different label refinement techniques.}
    \label{tab:ablation_study2}
    \vspace{-3.5mm}
    \end{table}

    \begin{figure}
        \centering
        \begin{subfigure}[b]{0.49\columnwidth}
            \centering
            \includegraphics[width=\linewidth]{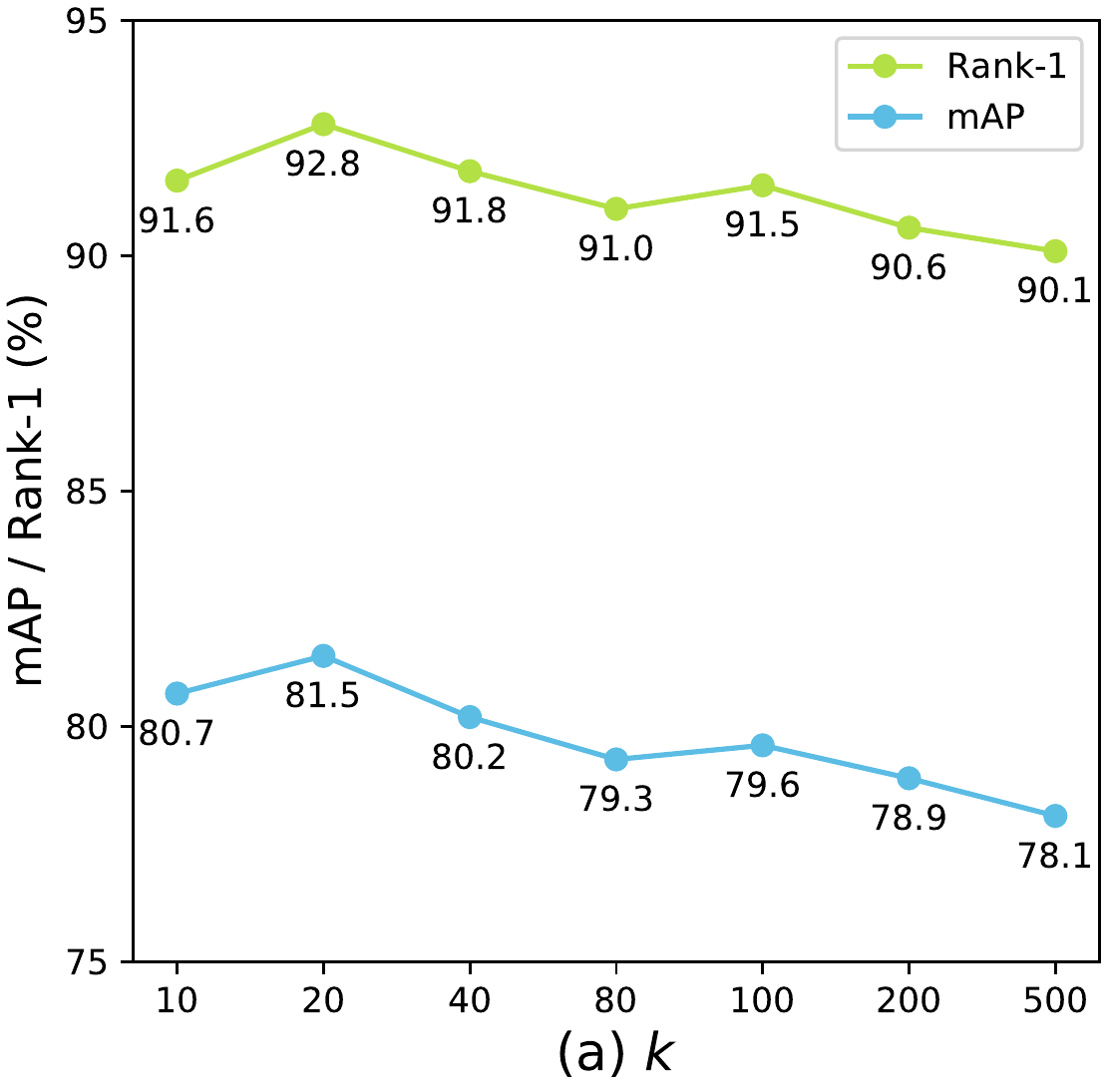}
            \label{fig:k_analysis}
        \end{subfigure}
        \hfill
        \begin{subfigure}[b]{0.49\columnwidth}
            \centering
            \includegraphics[width=\linewidth]{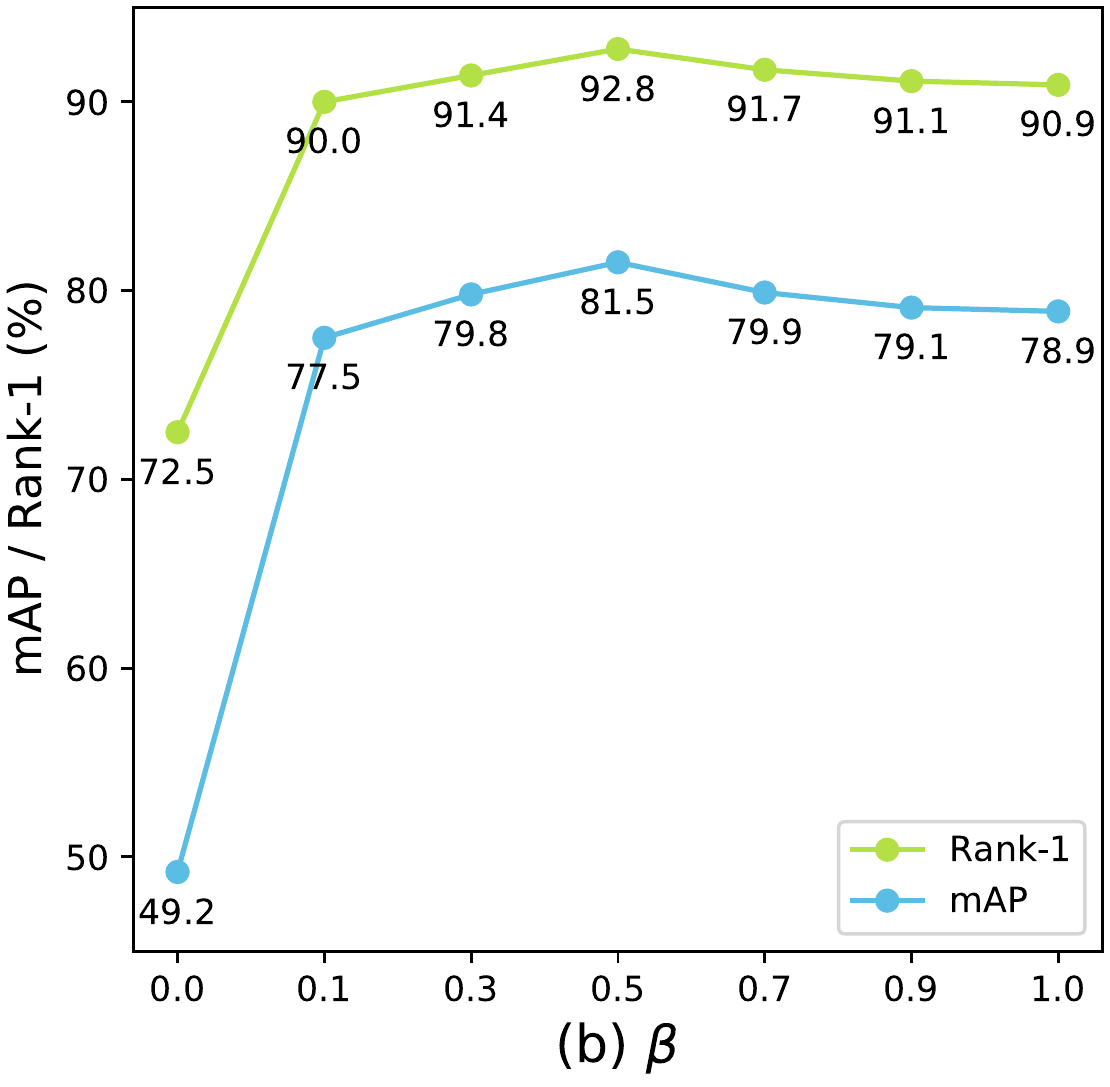}
            \label{fig:beta_analysis}
        \end{subfigure}
        \vspace{-5mm}
        \caption{Parameter analysis of $k$ and $\beta$ on Market-1501.}
    \vspace{-5.5mm}
    \label{fig:parameter_analysis}
    \end{figure}

\paragraphTitle{Effectiveness of agreement-aware label smoothing.}{
    To verify the necessity of AALS, we evaluate label refinement techniques alternative to AALS. 
    We first adopt vanilla label smoothing that employs a constant smoothing weight in Eq.~\eqref{eq:ls}.
    We also explore part-to-part label refinement, which generates refined labels using the predictions of other parts in the same way as PGLR in Eq.~\eqref{eq:pglr}.
    Specifically, we generate refined labels for the part $p_i$ by aggregating the predictions of the other parts $p_j$ with part-to-part cross agreement scores $\mathcal{C}_i(p_i, p_j)$.
    As shown in Table~\ref{tab:ablation_study2}, our AALS significantly outperforms other label refinement methods. 
    We observe that the part-to-part cross agreement score is substantially lower than the part-to-global counterpart because the parts share less overlapped receptive fields.
    Thus, part-to-part label refinement is guided by unreliable information from significantly different local contexts, while our AALS captures reliable complementary information and shows better performance.
    Vanilla label smoothing adjusts the label distribution for all parts blindly without considering the characteristics of each part and also shows limited improvement.
    Meanwhile, our AALS calibrates the part features' predictions according to the reliability of given labels captured by the cross agreements, and the well-calibrated part predictions lead to more reliable part-guided refined labels.
}

\paragraphTitle{Effectiveness of part-guided label refinement.}{
    To verify the effectiveness of PGLR, we evaluate other label refinement techniques. 
    One way is to refine labels with the prediction of global features by the mean-teacher model~\cite{tarvainen2017mean}.
    We further investigate PGLR without cross agreement scores by averaging the predictions of part features, \textit{i.e}, set all $w^{p_n}_i$ to $1/N_p$ in Eq.~\eqref{eq:pglr}.
    As shown in Table~\ref{tab:ablation_study2}, our PGLR significantly outperforms other label refinement methods.
    It demonstrates the superiority of PGLR and the effectiveness of the cross agreement score.
    The refined pseudo-label by PGLR captures reliable fine-grained information that cannot be achieved by considering only global features, and it helps to generate more effective refined labels.
}

\paragraphTitle{Parameter analysis.}{
    We analyze the impact of two parameters in our method -- the number of $k$-nearest neighbors for cross agreement scores and the weighting parameter $\beta$ for the part-guided label refinement.
    We tune the value of each parameter while keeping the others fixed, and the results are in Fig.~\ref{fig:parameter_analysis}.
    Large $k$ values result in more frequent false matches in top-$k$ ranked lists of global and part features, producing lower cross agreement scores overall.
    These false matches prevent us from identifying the reliability of complementary relationships, thus limiting performance.
    When we set $\beta$ to 0, our method decomposes down to using only the ensembled part predictions in Eq.~\eqref{eq:pglr}, showing the significant performance drop.
    The predictions of the initial training stage usually output uniform distributions, so the labels refined by PGLR also collapse to uniform distributions, providing noisy training signals.
    Based on these experimental results, we set $k=20$ and $\beta=0.5$.
}

    \begin{figure}[t]
    \centering
    \includegraphics[width=\columnwidth]{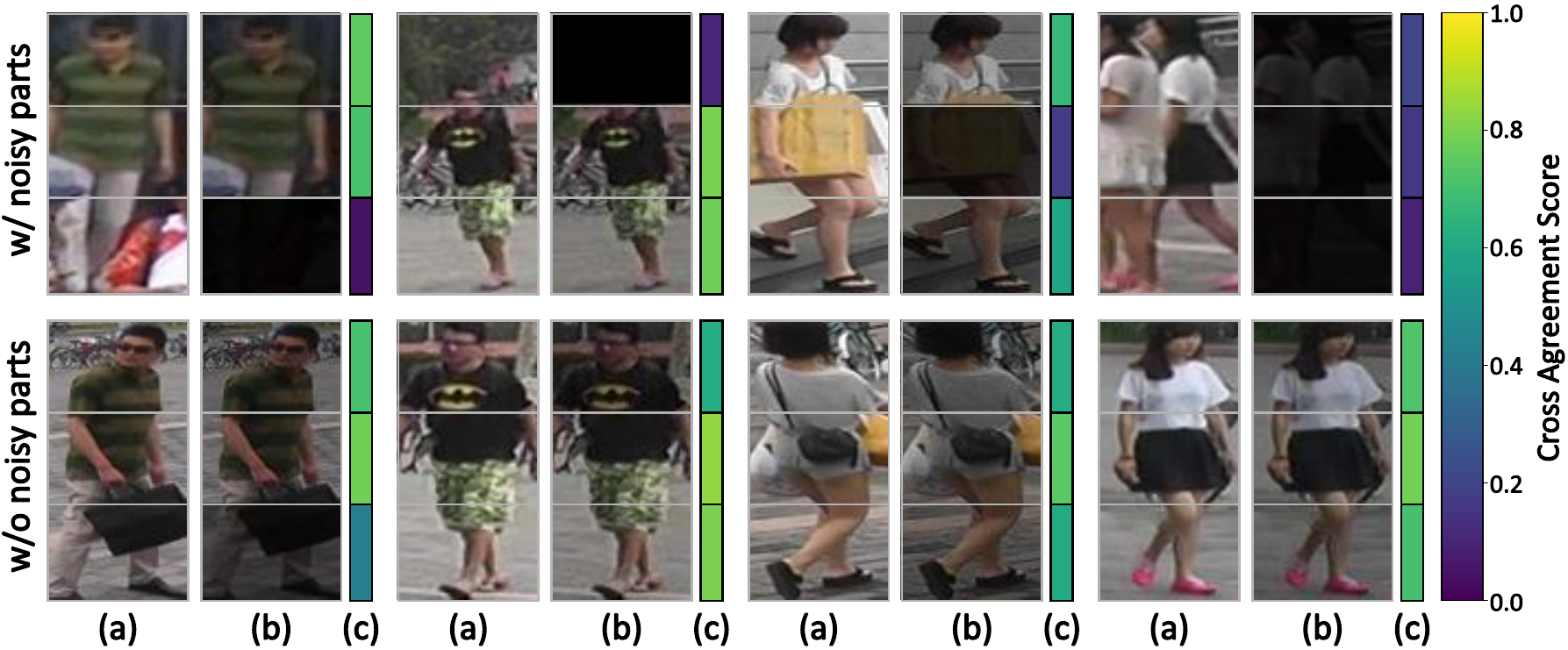}
        \caption{
        Visualization of cross agreement scores.
        (a) Original images.
        (b) Masked images created by multiplying the image intensity with the corresponding cross agreement score.
        (c) The color jet bar of the cross agreement score, where the color of each region indicates the average score over the entire training epochs.
        }
    \vspace{-5mm}
    \label{fig:qual}	
    \end{figure}

\paragraphTitle{Qualitative analysis.}{
    To explore the effect of cross agreement scores, we visualize the masked image $I^\prime(x, y) = I(x, y) \circ c(x, y)$ as the element-wise product between the intensity of the image $I(x, y)$ and the corresponding cross agreement score $c(x, y)$.
    A low cross agreement score leads to a low pixel intensity; thus, the corresponding part of an image becomes dark.
    As shown in Fig.~\ref{fig:qual}, occluded or misaligned parts have relatively low cross agreement scores while discriminative parts have high values.
    In the rightmost case on the top row of Fig.~\ref{fig:qual}, all parts fail to capture discriminative information due to occlusions by several people, and cross agreement values for all three parts are low, collapsing predictions of all parts to a uniform distribution by Eq.~\eqref{eq:aals}.
    The ensembled prediction of part-guided refined labels in Eq.~\eqref{eq:pglr} also collapses to a uniform vector, providing meaningless training signals.

    \begin{figure}[t]
    \centering
    \includegraphics[width=\columnwidth]{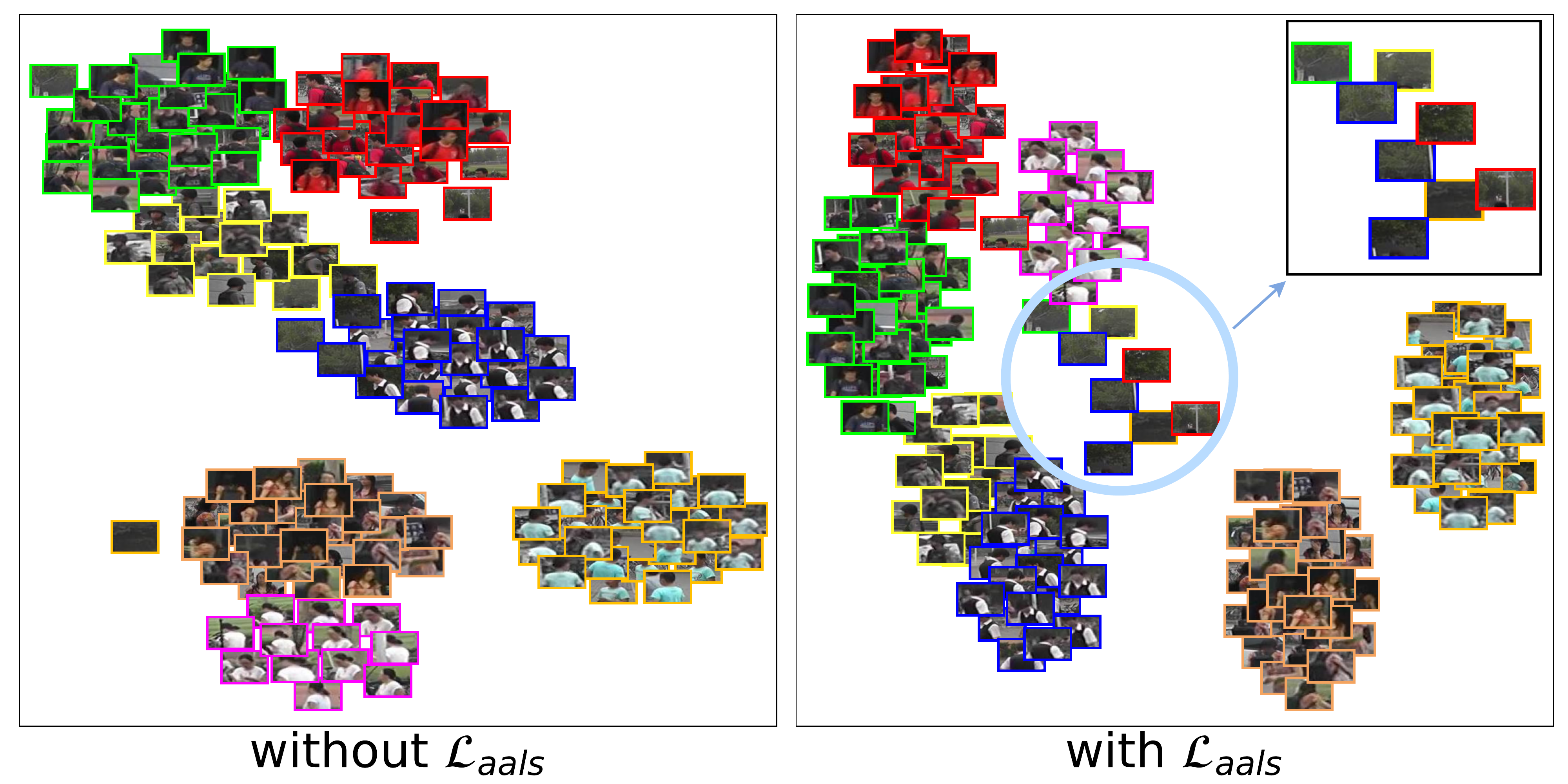}
        \caption{
        The t-SNE~\cite{van2008visualizing} visualization of the topmost part feature space of our PPLR without and with AALS.
        Without AALS, the embeddings overfit to ID labels and consider less the local context of each part.
        With AALS, it shows reliable embedding results that are well-distributed while considering the local context.
        The circled region shows that the features with less discriminative parts are embedded together even though their ID labels are different.
        }
    \vspace{-3mm}
    \label{fig:tsne_aals}
    \end{figure}

    \begin{figure}[t]
    \centering
    \includegraphics[width=\columnwidth]{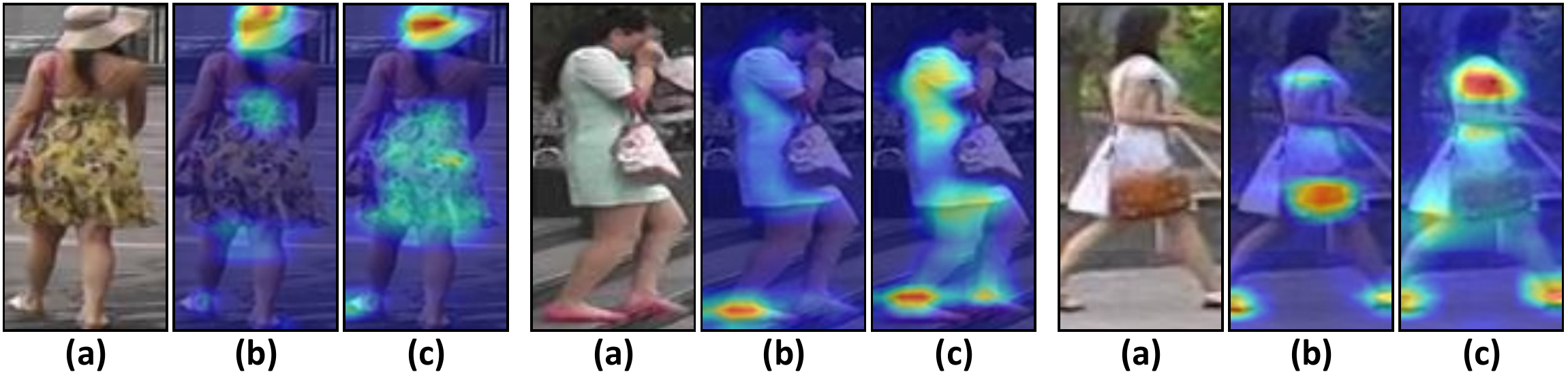}
        \caption{
        Grad-CAM~\cite{selvaraju2017grad} visualization of the global features' predictions: (a) Original images;
        (b) without $\mathcal{L}_{pglr}$; (c) with $\mathcal{L}_{pglr}$.
        (b) Without PGLR, the model focuses only on the most discriminative parts in the global context.
        (c) Thanks to the part-guided refined labels with rich local contexts, the model focuses on more diverse discriminative regions.
        }
    \vspace{-5mm}
    \label{fig:cam}
    \end{figure}
    
    \begin{table*}[!htbp]
    	\centering	
    	\begin{adjustbox}{max width=\textwidth}
    	\footnotesize
    	\begin{tabular}{l|c|cccc|cccc|cccc}
    	\hline
    		\multicolumn{2}{c|}{\multirow{2}{*}{Method}} & \multicolumn{4}{c|}{Market-1501} & \multicolumn{4}{c|}{MSMT17} & \multicolumn{4}{c}{VeRi-776} \\ \cline{3-14} 
     		\multicolumn{2}{c|}{} & mAP & R1 & R5 & R10 & mAP & R1 & R5 & R10 & mAP & R1 & R5 & R10 \\\hline
    		\multicolumn{14}{l}{\textit{Unsupervised methods without any labels}} \\ \hline
    		BUC~\cite{lin2019aBottom} & AAAI'19   & 38.3 & 66.2 & 79.6 & 84.5 & - & - & - & - & - & - & - & - \\
    		MMCL~\cite{wang2020unsupervised} & CVPR'20  & 45.5 & 80.3 & 89.4 & 92.3 & 11.2 & 35.4 & 44.8 & 49.8 & - & - & - & - \\
    		HCT~\cite{zeng2020hierarchical} & CVPR'20   & 56.4 & 80.0 & 91.6 & 95.2 & - & - & - & - & - & - & - & - \\
    		MMT~\cite{ge2020mutual} & ICLR'20 & 74.3 & 88.1 & 96.0 & 97.5 & - & - & - & - & - & - & - & -\\
    		GCL~\cite{chen2021joint} & CVPR'21 & 66.8 & 87.3 & 93.5 & 95.5 & 21.3 & 45.7 & 58.6 & 64.5 & - & - & - & -\\
    		SpCL~\cite{ge2020selfpaced} & NeurIPS'20 & 73.1 & 88.1 & 95.1 & 97.0 & 19.1 & 42.3 & 56.5 & 68.4 & 36.9 & 79.9 & 86.8 & 89.9 \\
    		RLCC~\cite{zhang2021refining} & CVPR'21 & {77.7} & {90.8} & {96.3} & {97.5} & {27.9} & {56.5} & 68.4 & 73.1 & 39.6 & 83.4 & 88.8 & 90.9 \\
    		\textbf{PPLR (Ours)} & This work 
    		& \textbf{81.5} & \textbf{92.8} & \textbf{97.1} & \textbf{98.1} 
            & \textbf{31.4} & \textbf{61.1} & \textbf{73.4} & \textbf{77.8}
            & \textbf{41.6} & \textbf{85.6} & \textbf{91.1} & \textbf{93.4} \\
    		\hline
    		\multicolumn{14}{l}{\textit{Unsupervised methods using camera labels}} \\ \hline
    		SSL~\cite{lin2020unsupervised} & CVPR'19   & 37.8 & 71.7 & 83.8 & 87.4 & - & - & - & - & - & - & - & -\\
    		JVTC~\cite{li2020joint} & ECCV'20 & 47.5 & 79.5 & 89.2 & 91.9 & 17.3 & 43.1 & 53.8 & 59.4 & - & - & - & - \\
    		IICS~\cite{xuan2021intra} & CVPR'21 & 72.9 & 89.5 & 95.2 & 97.0 & 26.9 & 56.4 & {68.8} & {73.4} & - & - & - & -\\
    		CAP~\cite{wang2021camera} & AAAI'21 & 79.2 & 91.4 & 96.3 & 97.7 & {36.9} & {67.4} & {78.0} & {81.4} & - & - & - & -\\
    		ICE~\cite{chen2021ice} & ICCV'21
    		& {82.3} & {93.8} & {97.6} & {98.4}
    		& {38.9} & {70.2} & {80.5} & {84.4}
    		& - & - & - & -\\
    		\textbf{PPLR (Ours)} & This work 
    		& \textbf{84.4} & \textbf{94.3} & \textbf{97.8} & \textbf{98.6}
            & \textbf{42.2} & \textbf{73.3} & \textbf{83.5} & \textbf{86.5}
            & \textbf{43.5} & \textbf{88.3} & \textbf{92.7} & \textbf{94.4} \\
    	\hline	
    	\end{tabular}
    	\end{adjustbox}
    	\caption{Comparison with the state-of-the-art methods on Market1501, MSMT17, and VeRi-776.}
    	\vspace{-5mm}
    \label{tab:sota_comparison}
    \end{table*} 

    To further explore the proposed label refinement methods, we qualitatively analyze the effectiveness of AALS and PGLR.
    We visualize the embeddings of the part feature of our PPLR with and without AALS.
    As shown in Fig.~\ref{fig:tsne_aals}, the embeddings without AALS are overfitted to the ID labels and show less discrimination over less informative parts.
    On the other hand, embeddings with AALS consider the semantic context of each part and are better distributed.
    While the model without AALS relies on hard pseudo-labels, AALS adjusts the given pseudo-labels based on the reliability of each part and achieves more proper part representation learning with a calibration effect.

    We also visualize Grad-CAM~\cite{selvaraju2017grad} of the predictions by global features with and without PGLR.
    As shown in Fig.~\ref{fig:cam}, the former focuses only on the most discriminative parts in the global context (\textit{e.g.}, hats, shoes, and bags) and is vulnerable to viewpoint changes and occlusions.
    Thanks to PGLR, which refines global pseudo-labels with rich fine-grained information, our model learns discriminative information from each part and captures diverse regions.
}

\subsection{Comparison with State-of-the-Arts}
    We compare our method with state-of-the-art unsupervised re-ID methods on Market-1501, MSMT17, and VeRi-776, and all the results are in Table~\ref{tab:sota_comparison}.

    We first compare our PPLR without the inter-camera contrastive loss (\textit{i.e.}, $\lambda_{cam}=0$) with unsupervised methods that do not use camera labels: BUC~\cite{lin2019aBottom}, MMCL~\cite{wang2020unsupervised}, HCT~\cite{zeng2020hierarchical}, MMT~\cite{ge2020mutual}, SpCL~\cite{ge2020selfpaced}, GCL~\cite{chen2021joint}, and RLCC~\cite{zhang2021refining}.
    SpCL employs a robust clustering criterion by identifying unreliable clusters to simulate accurate pseudo labels.
    MMT and RLCC refine noisy pseudo-labels using the predictions of an auxiliary network and a cluster consensus matrix, respectively.
    However, all these methods consider only global features and neglect the fine-grained information essential to person re-ID.
    On the other hand, our method considers the complementary relationship between global and part features and collectively mitigates the label noise for global features through the ensemble of the part predictions.
    As shown in Table.~\ref{tab:sota_comparison}, our method surpasses prior state-of-the-art methods on all the benchmarks with +3.8\%, +3.5\%, and +2.0\% of mAP than RLCC on Market-1501, MSMT17, and VeRi-776, respectively.

    We also compare our PPLR with unsupervised methods trained with camera labels: SSL~\cite{lin2020unsupervised}, JVTC~\cite{li2020joint}, IICS~\cite{xuan2021intra}, CAP~\cite{wang2021camera}, and ICE~\cite{chen2021ice}.
    IICS and CAP exploit intra- and inter-camera similarities and train the model based on each similarity to reduce large intra-class variance from different camera views.
    ICE leverages inter-instance pairwise similarities based on data augmentation strategies to boost contrastive learning schemes.
    Our method also utilizes various feature similarity structures, but we focus on exploiting both the global and local similarity information desirable for fine-grained person re-ID.
    Additionally, unlike the other methods that directly use pseudo-labels computed by feature similarities, our method considers the complementary information between the feature similarity structures.
    As shown in Table.~\ref{tab:sota_comparison}, our method significantly outperforms state-of-the-art methods by a remarkable margin;
    PPLR scores higher in mAP than ICE by +2.1\% and +3.3\% on Market-1501 and MSMT17, respectively.

\section{Discussion}
    While we have shown the effectiveness of our method, it still has a limitation to overcome.
    Like other part-based methods~\cite{sun2018beyond, zheng2019pyramidal, fu2019self}, we extract part features by partitioning a feature map uniformly.
    Since not all images are aligned to a human body, however, part features in the same part index may not represent the same body parts. 
    Even though they capture discriminative information, misaligned part features may output ranked lists with noisy information and produce low cross agreement scores. 
    To overcome this, a promising solution would be using semantic matching or human parsing techniques to construct feature spaces that represent a shared, similar semantic part of a person.
    
    In addition, the re-ID techniques may potentially bring negative impacts, such as infringement of privacy due to abuse of a surveillance system.
    Related researchers and users should be attentive to using the technology in an appropriate manner while considering ethical issues.
    DukeMTMC-reID~\cite{ristani2016performance} has been taken down due to ethical concerns and should no longer be used. \vspace{-0.5mm}

\section{Conclusion}
\label{sec:5}
In this paper, we have proposed a Part-based Pseudo Label Refinement (PPLR) framework for unsupervised person re-identification. Our method exploits both the global and local context of an image and alleviates the label noise for each feature space using the complementary relationship between the global and part features.
We have introduced a cross agreement score to leverage reliable complementary information, and based on this, we have proposed agreement-aware label smoothing and part-guided label refinement.
We have conducted extensive ablation studies, including qualitative analysis, to validate the effectiveness of the proposed method.
Furthermore, extensive experiments have shown that our framework outperforms prior state-of-the-art methods on several benchmarks.

{\small\paragraphTitle{Acknowledgement}{
    This work was supported in part by the Institute of Information \& communications Technology Planning \& Evaluation (IITP) grant funded by the Korea government (MSIT) (No. IITP-2015-0-00199, Proximity computing and its applications to autonomous vehicle, image search, and 3D printing) and the National Research Foundation of Korea (NRF) grant funded by the Korea government (MSIT) (No. NRF-2021R1C1C1012540).}
}

{\small
\bibliographystyle{bib/ieee_fullname}
\bibliography{bib/egbib}
}

\clearpage
\appendix
\noindent{\Large{\textbf{Appendix}}}
\section{More Experimental Results}
\label{sec:experiment-supp}
\subsection{Parameter Analysis}
\begin{figure}[t]
    \centering
    \begin{subfigure}{\columnwidth}
      \centering
      \includegraphics[width=0.47\columnwidth]{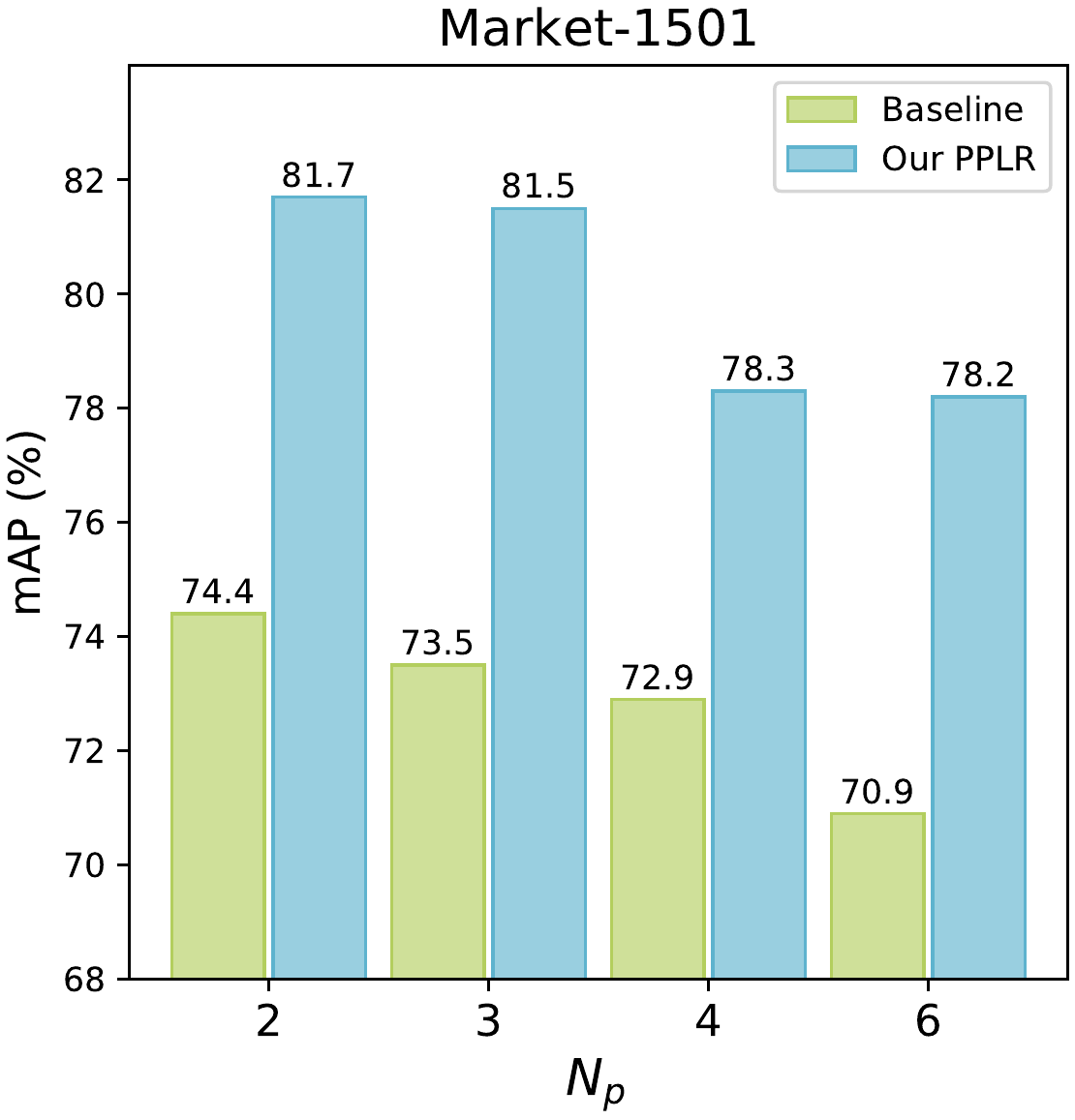} 
      \hfill
      \includegraphics[width=0.47\columnwidth]{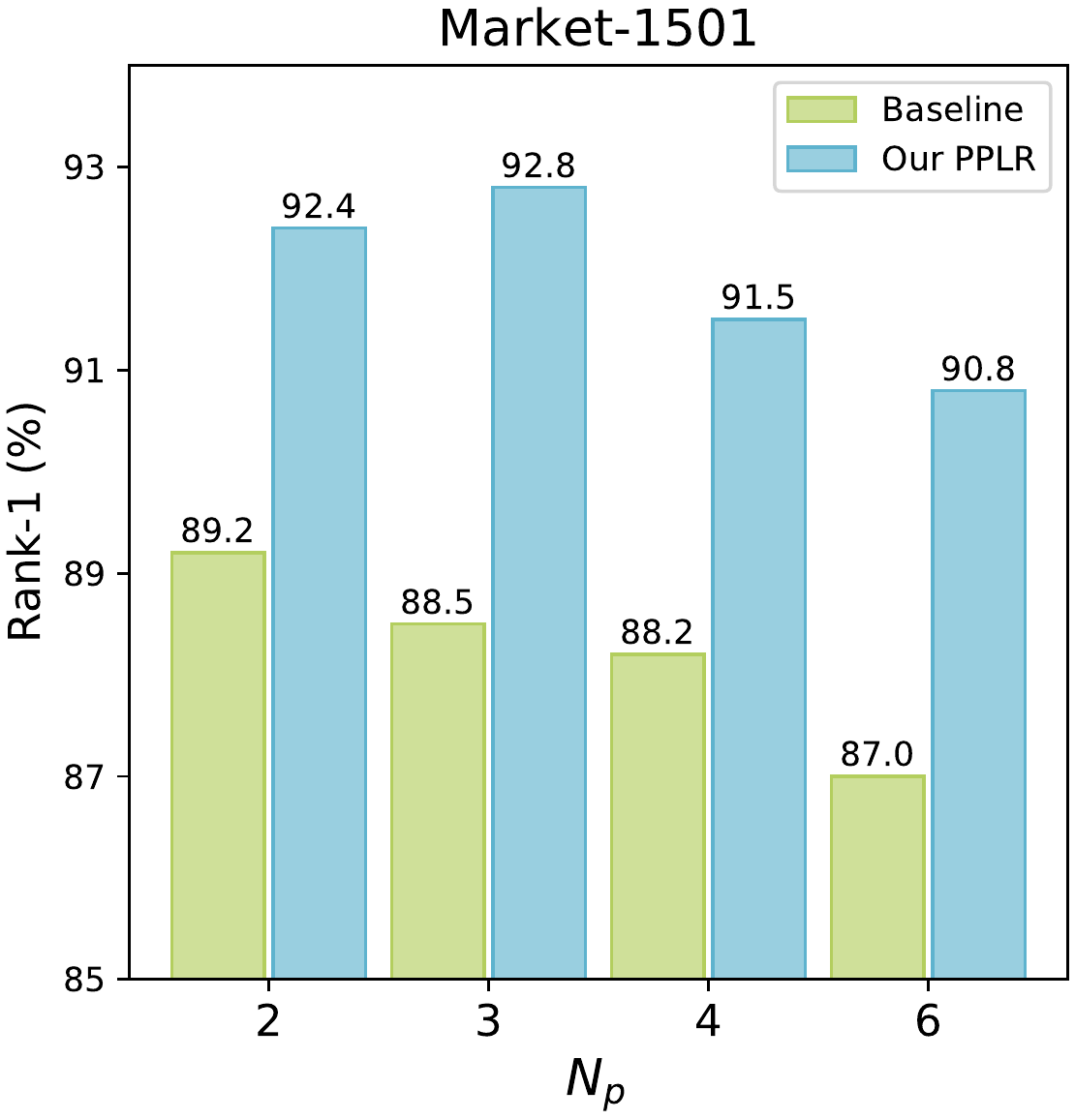} 
      \label{fig:n_analysis_market}       
    \end{subfigure}
    \caption{
    Parameter analysis of $N_p$ on Market-1501.
    `Baseline' is the part-based unsupervised re-ID framework (Sec.~\ref{ssec:3-1}).
    }
    \label{fig:n_analysis}  
    \end{figure}

{
    We analyze the impact of the number of parts $N_p$ in our part-based framework, and the results are in Fig.~\ref{fig:n_analysis}.
    Larger values of $N_p$ reduce the receptive field of the part feature to contain limited clues for re-identifying a person.
    Thus, the baseline performance is decreased as the number of parts increases because the part features with smaller receptive fields are simply trained by hard pseudo-labels that do not consider the context of each part.
    Nevertheless, PPLR consistently improves the baseline performance with a significant margin throughout different values of $N_p$.
    Thanks to cross agreement scores, agreement-aware label smoothing adjusts the label distribution while considering the context of each part, leading to proper part feature learning.
}

\subsection{Training Computational Cost}
{
    We compare training computation costs of PPLR with other methods that utilize auxiliary teacher networks to refine pseudo-labels: MEB-Net\footnote{\url{https://github.com/YunpengZhai/MEB-Net}}~\cite{zhai2020multiple} and MMT\footnote{\url{https://github.com/yxgeee/MMT}}~\cite{ge2020mutual}.
    We analyze the number of training parameters, training stage time, and clustering stage time of each method, and the results are in Tab.~\ref{tab:cost}.
    PPLR only uses features from a single backbone, and it is more efficient than other methods, even including the time to compute the cross agreement score.
    On the other hand, MMT and MEB-Net use an averaged feature for each sample from multiple backbones for clustering, which requires additional computational cost in the clustering stage.
    While other methods leverage multiple networks (\textit{e.g.}, dual ResNet in MMT, and single DenseNet, ResNet, and Inception-v3 in MEB-Net),
    our PPLR is a self-teaching method and requires fewer parameters for training.
    Furthermore, other methods require multiple feedforwards to refine the pseudo-labels; \textit{e.g.,} MMT feedforwards two current models and two mean-teacher models a total of four times. 
    PPLR only requires a single feedforward for pseudo-label refinements and shows efficiency in the training stage.
}

\subsection{Qualitative Results}
\begin{figure*}[t]
        \centering
        \includegraphics[width=0.9\textwidth]{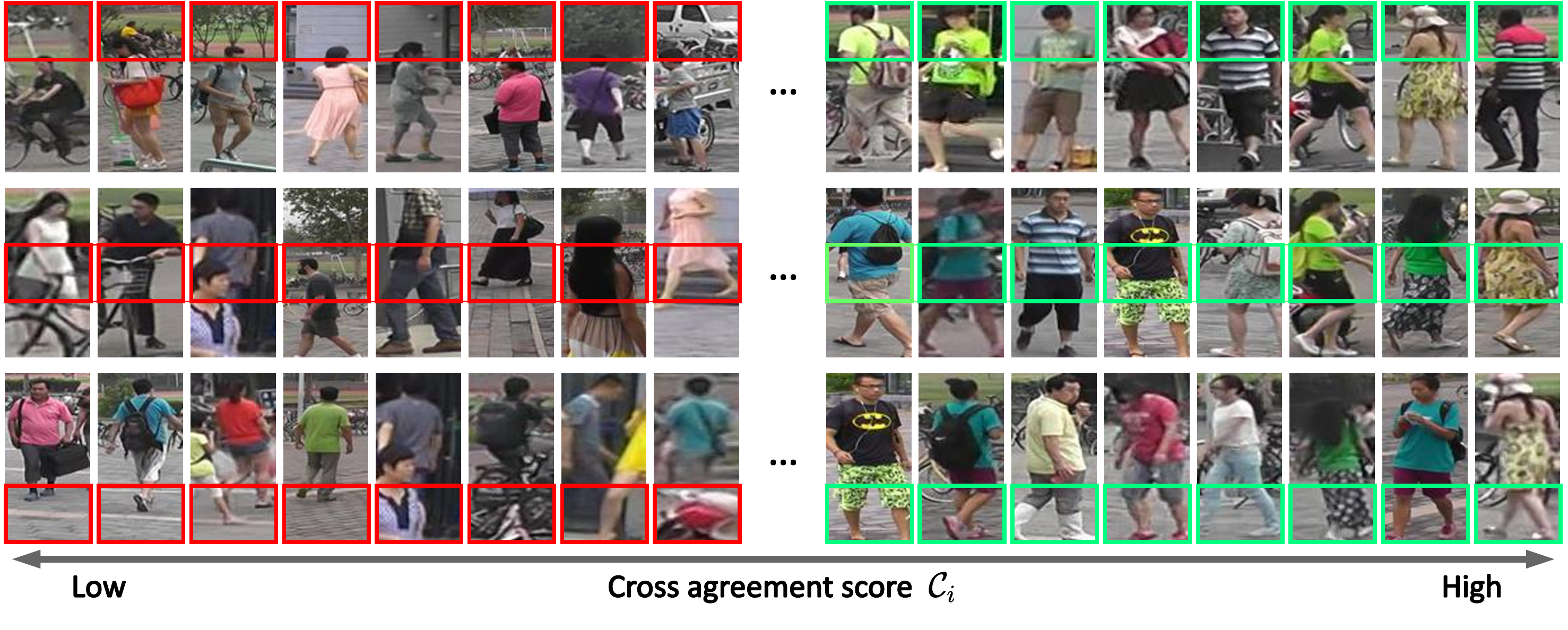}
        \caption{ 
        Visualization of images with low-50 and top-50 cross agreement scores on Market-1501. 
        Very similar and duplicated images were excluded to show various cases.
        }
        \vspace{-3mm}
        \label{fig:ca-market}
    \end{figure*}

{
    To further analyze cross agreement scores, we visualize images that have low- and top-50 cross agreement scores on Market-1501.
    As shown in Fig.~\ref{fig:ca-market}, the images with low cross agreement scores contain less discriminative information irrelevant to identifying a person in corresponding part (\textit{e.g.,} occlusions, backgrounds, and presence of multiple people).
    In contrast, the images with high cross agreement scores are well-aligned with discriminative information.
    There are also some failure cases to overcome that we leave for future work.
    Some misaligned parts with discriminative information have low cross agreement scores because they capture different body features compared to the corresponding parts in the rest of the images.
    To overcome this limitation, a promising solution would be using auxiliary human semantic information by person attribute recognition or human parsing techniques to construct feature spaces that represent similar semantic parts.
}

    \vspace{-3mm}
    
    \begin{table}[t]
    \centering
    \begin{adjustbox}{max width=\columnwidth}
    \begin{tabular}{c | c | c| c}
        	\hline
    		Method & Parameters (M) & \begin{tabular}{@{}c@{}}Clustering stage time \\ (sec / epoch) \end{tabular} & \begin{tabular}{@{}c@{}}Training stage time\\   (sec / iter) \end{tabular} \\
        	\hline
    		MEB-Net~\cite{zhai2020multiple} & 56.867 & 120.731 & 0.998  \\
    		MMT~\cite{ge2020mutual} & 50.096 & 48.545 & 0.511 \\
    		\hline
    		Baseline & 29.668 & 36.103 & 0.194 \\
    		PPLR & 29.668 & 38.484 & 0.202 \\
        	\hline
    \end{tabular}
    \end{adjustbox}
    \caption{
    Training cost comparison on Market-1501.
    The clustering stage time includes the time for feature extraction, clustering, and cross agreement score computation.
    Since the number of iterations per epoch is different for each method, we measure `sec/iter' for a fair evaluation of training stage time.
    }
    \vspace{-5mm}
    \label{tab:cost}
    \end{table}

\section{Camera-aware Proxy Details}
\label{sec:cam_details}
    When the camera labels are available, PPLR can optionally leverage the inter-camera contrastive loss with camera-aware proxies~\cite{wang2021camera}. 
    Let $y_i$ and $c_i$ respectively denote the pseudo-label and the camera label of the image $x_i$.
    We compute the camera-aware proxy $\mathbf{c}_{(a,b)}$, which is the centroid of the features $\mathbf{f}_i$ that have the same camera label $a$ and belong to the same cluster $b$, defined by:
    \begin{equation}
    \mathbf{c}_{(a,b)} = \frac{1}{\lvert S_{(a, b)} \rvert} \sum_{i \in S_{(a, b)}} \mathbf{f}_i,
    \label{eq:proxy}
    \end{equation}
    where $S_{(a, b)} = \{ i \lvert c_i=a \wedge y_i=b\}$ is the index set for the proxy $\mathbf{c}_{(a,b)}$, and $\lvert \cdot \rvert$ is the cardinality of the set.
    
    To compute the inter-camera contrastive loss, the index set $\mathcal{P}_{i}$ for the positive proxies of the feature $\mathbf{f}_{i}$ is defined as the proxy indices that have the same pseudo-label $y_i$ but different camera labels with $\mathbf{f}_{i}$.
    The index set $\mathcal{Q}_i$ for the hard negative proxies of the feature $\mathbf{f}_{i}$ is defined as the indices of nearest proxies that have different pseudo-labels to $y_i$.
    We utilize the inter-camera contrastive loss with camera-aware proxies on each feature space to reduce the large intra-class variance by disjoint camera views.
    
\section{More Implementation Details}
\label{sec:implement_details}
    We implement our framework based on PyTorch.
    Four NVIDIA TITAN RTX GPUs are used for training, and only a single GPU is used for testing.
    We compute the Jaccard distance based on the $k$-reciprocal encoding~\cite{zhong2017re} for clutsering, where $k$ is set to 30.
    For parameters of DBSCAN~\cite{ester1996density}, we set the minimum number of neighbors for a core point to 4 and the distance threshold between samples to 0.7 for MSMT17 and VeRi-776 and 0.6 for Market-1501.
    With the inter-camera contrastive loss, we use smaller distance threshold between samples, \textit{e.g.}, 0.6 for MSMT17.
    For stable training, we apply the agreement-aware label smoothing after the first five epochs.
    % We apply agreement-aware label smoothing after the first five epochs.

\begin{figure*}[t]
    \centering
    \includegraphics[width=0.8\textwidth]{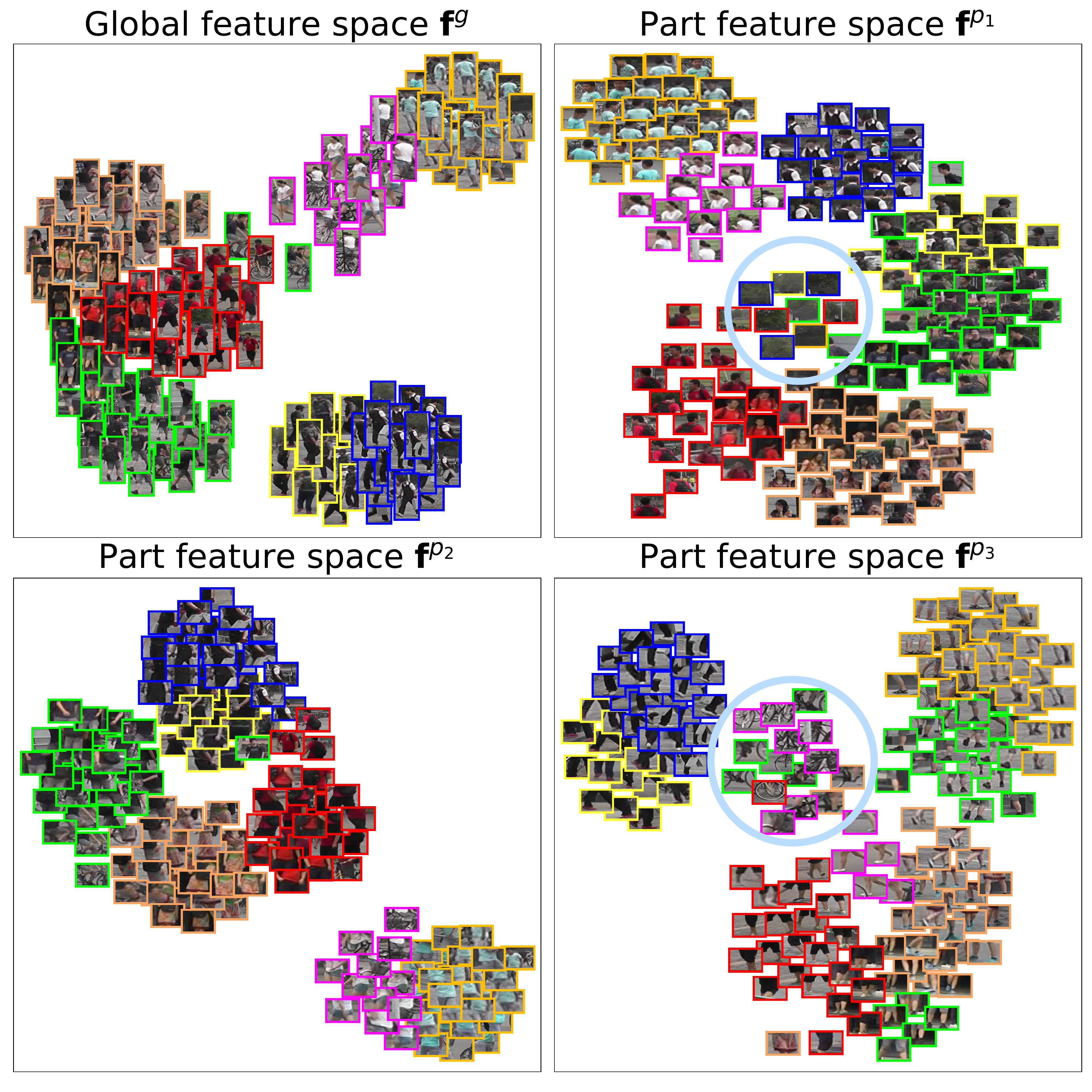}
    \caption{ 
    The large version of Fig.~\ref{fig:tsne-each-space}.
    }
    \label{fig:tsne-each-space-large}
\end{figure*}

\begin{figure*}[t]
    \centering
    \includegraphics[width=0.8\textwidth]{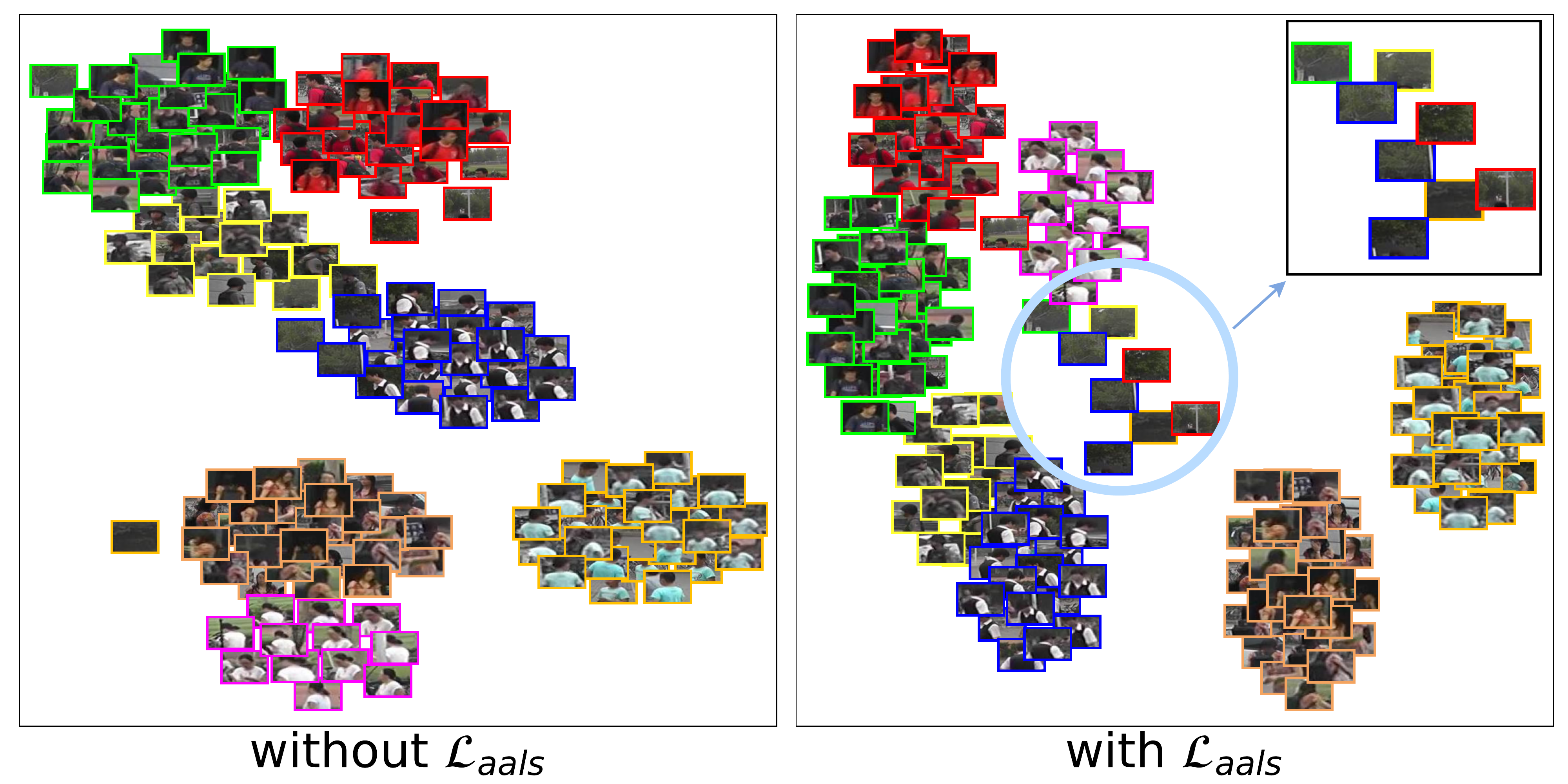}
    \caption{ 
    The large version of Fig.~\ref{fig:tsne_aals}.
    }
    \label{fig:tsne-aals-large}
\end{figure*}

\end{document}